%% file: main.tex
\documentclass[sigconf,nonacm]{acmart}

\AtBeginDocument{%
}

\renewcommand\footnotetextcopyrightpermission[1]{}

\usepackage{booktabs}
\usepackage{amsmath}
\usepackage{array}
\usepackage{graphicx}
\usepackage{multirow}
\usepackage{algorithm}
\usepackage{algorithmic}
\usepackage{xcolor}

\begin{document}
\title{Adaptive Finite-Budget Training for CVaR Risk-Aware Q-Learning}

\author{Yifan Wu}
\email{wuyifan0648@163.com}
\affiliation{%
  \department{Department of Data and Systems Engineering}
  \institution{The University of Hong Kong}
  \city{Hong Kong}
  \country{Hong Kong SAR}
}

\author{Junjie Lei}
\authornote{Primary contact.}
\email{junjie01@connect.hku.hk}
\affiliation{%
  \department{Department of Data and Systems Engineering}
  \institution{The University of Hong Kong}
  \city{Hong Kong}
  \country{Hong Kong SAR}
}

\author{Wenjie Huang}
\email{huangwj@hku.hk}
\affiliation{%
  \department{Department of Data and Systems Engineering \& Musketeers Foundation Institute of Data Science}
  \institution{The University of Hong Kong}
  \city{Hong Kong}
  \country{Hong Kong SAR}
}

\renewcommand{\shortauthors}{Wu, Lei, and Huang}

\begin{abstract}
Risk-aware Q-learning (RaQL) provides a model-free, two-timescale estimator for dynamic risk objectives, but its finite-budget behavior remains fragile: fixed inner-loop hyperparameters can produce unstable value estimates, persistent Bellman residuals, and inefficient sample reuse. This paper proposes an adaptive training controller for Conditional Value-at-Risk (CVaR) RaQL and evaluates it on a daily Bitcoin trading task. The controller preserves the original CVaR estimator and Bellman fixed point; instead, it redesigns the training procedure through six coordinated mechanisms: per-cell inner-step sizing, outer-rate-matched decay synchronization, a short early correction for the VaR-like inner variable, a coverage-first-then-greedy sample allocation rule, progressive suffix aggregation of mature inner estimates, and data-driven calibration of key scales from online-observable quantities. Across 20 random seeds and 856,000 inner-transition samples, the controller reduces the mean empirical CVaR Bellman residual by approximately 85\% relative to the fixed-parameter baseline (MeanBEQ: 1.2202 to 0.1854; MeanBEV: 1.1624 to 0.0535) and maintains stability across CVaR levels, discount factors, and training budgets. On the chronological out-of-sample test set, the learned policy attains a Sharpe ratio of 0.9281 with a maximum drawdown of 6.46\% after transaction costs. Although buy-and-hold yields a higher cumulative return (35.43\% vs. 23.61\%), the adaptive policy achieves far lower volatility (9.57\% vs. 47.93\%), drawdown, and CVaR loss. These results demonstrate that adaptive finite-budget training design, applied solely to the training procedure without altering the risk objective, can materially improve the reliability and risk-adjusted performance of risk-aware Q-learning in financial applications.
\end{abstract}

%\ccsdesc[500]{Computing methodologies~Reinforcement learning}
%\ccsdesc[300]{Applied computing~Economics}


\maketitle

\section{Introduction}
%\WJcomment{I prefer to merge Section 1 and 2, three things shall be contained:
%\begin{enumerate}
%    \item add more literature on risk+rl, position RAQL as the first paper give non-asymptotic convergence rate. (introduce cvar section 2.1 now somewhere)
%    \item present the sudo-algorithm of RAQL
%    \item criticize its sample complexity and slow convergence rate via numerical evidence and structural analysis. (include section 2.2

%You can do the first two things, and leave the third for me.
%\end{enumerate}
%}

\subsection{Risk-Sensitive Reinforcement Learning in Finance and RaQL}
Financial systems are often evaluated by downside exposure and tail losses rather than mean return alone. This has motivated risk-sensitive reinforcement-learning (RL) applications in algorithmic trading, foreign exchange, hedging, and portfolio allocation. Shen et al.~\citep{shen2014riskaverse} study risk-averse algorithmic trading using high-frequency NASDAQ data, while Bisi et al.~\citep{bisi2020fx} combine fitted Q-iteration with a risk--return objective for foreign-exchange trading. Vittori et al.~\citep{vittori2020hedging} consider option hedging under transaction costs, and Vadori et al.~\citep{vadori2020martingale}, Coache et al.~\citep{coache2023dynamic}, and Enkhsaikhan and Jo~\citep{enkhsaikhan2024portfolio} develop risk-aware formulations for portfolio optimization and statistical arbitrage. These studies introduce risk sensitivity through the reward objective, constraints, uncertainty representation, or policy architecture. %{\color{blue}{WJ: We need several papers for risk-averse RL in finance applications e.g., portfolio optimization. }}

Among the available risk measures, Conditional Value-at-Risk (CVaR) is particularly relevant because it averages losses in the worst $1-\alpha$ tail beyond the Value-at-Risk threshold. CVaR is coherent \citep{artzner1999,acerbi2002} and admits a tractable variational representation \citep{rockafellar2000}. In financial RL, CVaR has been incorporated into direct and deep trading methods \citep{alameer2022cvar,cui2023crypto}, while policy-gradient and actor--critic methods have incorporated CVaR as an objective or constraint \citep{chow2014,prashanth2014}. Distributional RL instead estimates return quantiles and extracts risk-sensitive policies from the learned return distribution \citep{dabney2018,lim2022,chen2024}.

Our starting point is the work \citet{huang2021}, which develops risk-aware Q-learning (RaQL) under dynamic coherent risk measures. RaQL replaces the conditional expectation in the Bellman recursion with a risk functional, estimates that functional through a stochastic approximation inner loop, and updates Q-values in an outer loop. For CVaR, the inner recursion tracks a VaR-like minimizer. As summarized by \citet{wang2022risk}, \citet{huang2021} are the first to establish almost-sure convergence and an explicit non-asymptotic rate for this model-free two-loop algorithm, building on classical Q-learning and two-timescale stochastic approximation \citep{watkins1992,borkar2000,konda2000}.

The theory specifies schedules under which the coupled recursions converge, but it does not prescribe how to allocate a fixed sampling budget among inner risk estimation, outer Q-updates, and state--action cells. The allocation will affect the risk estimation error and Q-updates error, and the two errors will be coupled during training. %These decisions interact: inner CVaR error changes the outer target, while each Q-update shifts the distribution that the inner recursion must track. Under CVaR, a VaR-like inner variable that lags behind its moving minimizer can further amplify the outer target.
Section 1.2 and 1.3 make this mechanism precise and motivate changing the training controller without changing the CVaR estimator or Bellman fixed point. 

\emph{Adaptive training} has been studied extensively for risk-neutral and single-timescale learning agents. It means that some part of the learning procedure is adjusted using the explored state--action pairs rather than fixed once before training. Existing work illustrates several forms of adaptation. Learning-rate and sample-efficiency analyses characterize how update schedules and exploration govern Q-learning efficiency \citep{evendar2003,jin2018}. Adaptive subgradient methods rescale coordinate-wise updates using accumulated gradient information \citep{duchi2011}, while adaptive value- and return-normalization methods update the scale of learning targets from running statistics \citep{vanhasselt2016,schaul2021}. In exploration, count-based methods reward rarely visited states through visitation-based bonuses \citep{bellemare2016}, whereas value-driven methods adjust the exploration probability according to observed value differences \citep{tokic2010}. Prioritized computation focuses updates on states expected to change most or on transitions with large temporal-difference errors \citep{moore1993,schaul2016}. Iterate averaging combines successive Q-updates to reduce stochastic variance \citep{li2023}, and nested-simulation theory allocates a finite simulation budget between outer scenarios and inner estimation \citep{gordy2010}. These mechanisms explain concretely what may be adapted: the step-size, target scale, exploration rule, update priority, estimator aggregation, or sample allocation. However, none directly resolves the two-timescale coupling in CVaR RaQL under a fixed budget. We therefore adapt the RaQL training controller while retaining its estimator and fixed point.

\subsection{CVaR Recursion and the RaQL Baseline}

This section represents the definition of CVaR and structure of RaQL Let $\mathcal S$ and $\mathcal A$ be the finite state and action sets, with $S:=|\mathcal S|$ and $A:=|\mathcal A|$, and let $Q:\mathcal S\times\mathcal A\to\mathbb R$ be the cost-valued action-value table. Let $X$ denote a random future loss. The upper-tail CVaR at level $\alpha\in(0,1)$ can be written in optimized-certainty-equivalent form as \citep{rockafellar2000,bental2007}
\begin{equation}
  \rho_{\alpha}(X)
  =
  \min_{y\in\mathbb{R}}
  \left\{
  y+\frac{1}{1-\alpha}\mathbb{E}\left[(X-y)_+\right]
  \right\},
  \label{eq:cvar-oce}
\end{equation}
where $(u)_+:=\max\{u,0\}$. For CVaR, the optimizer $y^\star$ is a Value-at-Risk threshold. For a state--action pair $(s,a)$ and sampled next state $s'$, define the state--action-dependent one-step loss target
\begin{equation}
  x_{sa}(s';Q) = \ell(s,a,s') + \gamma V_Q(s'), \quad
  V_Q(s)=\min_{a\in\mathcal{A}}Q(s,a),
  \label{eq:one-step-loss-target}
\end{equation}
where $\ell(s,a,s')$ is the immediate loss, $\gamma$ is the discount factor, and $s'$ is sampled from the unknown transition kernel. The algorithms abbreviate the realized $x_{sa}(s';Q)$ as $x$ and write the cell-specific threshold estimate as $y(s,a)$. For generic scalar arguments $x$ and $y$, the sample-level CVaR target and its subgradient are
\begin{align}
  G_{\alpha}(x,y)
  &=
  y+\frac{1}{1-\alpha}(x-y)_+ ,
  \label{eq:cvar-sample-target}\\
  g_y(x,y)
  &=
  1-\frac{1}{1-\alpha}\mathbf{1}\{x>y\}.
\end{align}
%\WJcomment{We shall add the state--action dependence on the $x$ and $y$. }
Accordingly, $x_{sa}(s';Q)$ and $y(s,a)$ are explicitly state--action dependent. The shorter symbols $x$ and $y$ are used only inside a sampled update. Algorithm \ref{alg:baseline-raql} gives the fixed-parameter RaQL implementation used as the reference baseline. %\WJcomment{What do you mean by Scheme 0 ?}
The total sampling budget $B$ is the number of inner-loop transition samples available for training: each sampled transition in the inner loop consumes one unit of $B$. %It is a computational sample limit, not a monetary trading budget. %\WJcomment{I do not understand what ``budget'' here mean}
The inner-loop depth $L$ is the maximum number of such transitions processed before the next outer Q-update. At the start of each outer iteration, $Q^{-}\gets Q$ creates a frozen copy of the current Q-table, and $V^{-}(s):=\min_a Q^{-}(s,a)$ is the next-state value computed from that copy. Both are held fixed during the inner loop. In addition to this within-iteration freezing, the inner-loop depth $L$ and the inner risk scale $h_y$ are both fixed throughout training, applies the within-loop decay factor $j^{-p}$, and retains only the final CVaR target for each visited cell. %\WJcomment{What do you mean by continuation table ?}
The counter $n_{sa}$ records the number of earlier outer Q-updates to cell $(s,a)$. Within an inner loop, $\mathcal{Y}_{sa}$ stores the iterates of $y(s,a)$, $\bar y_{sa}$ is the mean of the latter half of that queue, and $\widehat q_{sa}$ is the most recently retained CVaR target. The map $\Pi_{\mathcal Y}$ denotes projection onto the admissible interval $\mathcal Y$. %\WJcomment{In which experiments ???}
These fixed choices define the reference baseline, but they cannot adjust as training progresses. The next subsection explains why this can be problematic under a finite sampling budget. %\WJcomment{still do not understand.}

\begin{algorithm}[t]
\caption{Fixed-parameter CVaR RaQL baseline}
\label{alg:baseline-raql}
\begin{algorithmic}[1]
\REQUIRE Total inner-transition sample budget $B$; inner-loop depth $L$; base scale $h_y$; CVaR level $\alpha$; discount $\gamma$; inner and outer exponents $p,\eta$; admissible interval $\mathcal{Y}$ for $y$
\STATE Initialize $Q(s,a)\gets0$, $y(s,a)\gets0$, $n_{sa}\gets1$, and consumed-sample count $b\gets0$
\WHILE{$b<B$}
  \STATE Freeze $Q^{-}\gets Q$ and set $V^{-}(s)\gets\min_a Q^{-}(s,a)$
  \STATE Observe $s$ and choose action $a$ by exploration
  \STATE Reset per-cell queues $\mathcal{Y}_{sa}$ and retained targets $\widehat q_{sa}$
  \FOR{$j=1,\ldots,\min\{L,B-b\}$}
    \STATE Observe loss $\ell$ and next state $s'$; set $b\gets b+1$
    \STATE Append $y(s,a)$ to $\mathcal{Y}_{sa}$ and set $\bar y_{sa}\gets\operatorname{HalfWindowMean}(\mathcal{Y}_{sa})$
    \STATE $x\gets\ell+\gamma V^{-}(s')$, $q\gets G_\alpha(x,\bar y_{sa})$, and $\widehat q_{sa}\gets q$
    \STATE $y(s,a)\gets\Pi_{\mathcal Y}[y(s,a)-j^{-p}h_y g_y(x,\bar y_{sa})]$
    \STATE $s\gets s'$
  \ENDFOR
  \FORALL{$(s,a)$ visited in this outer iteration}
    \STATE $Q(s,a)\gets(1-n_{sa}^{-\eta})Q^{-}(s,a)+n_{sa}^{-\eta}\widehat q_{sa}$
    \STATE $n_{sa}\gets n_{sa}+1$
  \ENDFOR
\ENDWHILE
\RETURN $Q$ and its greedy policy
\end{algorithmic}
\end{algorithm}

\subsection{Finite-Budget Failure Mechanism and Research Gap}

The convergence guarantee of RaQL specifies conditions under which the risk estimates and Q-values approach the optimal solutions as the sampling continues, but it does not give a practical rule for allocating a fixed number of samples. %\WJcomment{I cannot catch the meaning of these terminologies.}
%A practitioner must still choose the number of inner samples between Q-updates and the scale of the threshold updates. 
The allocation may need to change across training and across state--action cells. This issue is especially important for CVaR RaQL because, when $y<x$,
\begin{equation}
  G_{\alpha}(x,y)
  =
  \frac{x}{1-\alpha}
  -
  \frac{\alpha y}{1-\alpha}.
  \label{eq:amplification}
\end{equation}
At $\alpha=0.8$, for example, $G_{\alpha}(x,y)=5x-4y$. Thus, if $y$ remains below the relevant VaR threshold, its error is amplified in the target passed to the outer Q-update rather than remaining confined to the inner recursion.

For a frozen table $Q^{-}$, define
$X_{sa}(Q^{-})=\ell(s,a,S')+\gamma V_{Q^{-}}(S')$, where $S^\prime$ is the next random state, and the corresponding CVaR optimizer
\[
y^\star_{sa}(Q^{-})
\in
\arg\min_y
\mathbb{E}\!\left[
y+\frac{(X_{sa}(Q^{-})-y)_+}{1-\alpha}
\mid s,a
\right].
\]
This optimizer is a conditional $\alpha$-VaR threshold. A \emph{lagging} inner variable means that $y(s,a)$ remains materially below the current optimizer $y^\star_{sa}(Q^{-})$; it does not mean that $y(s,a)$ should match an individual sampled target.

With a finite budget, the baseline fixes the inner-loop length and update scale before training. Inner estimation error then affects the outer update, while changes in the Q-table alter the target that the inner loop must subsequently estimate. A single fixed setting may therefore be unstable early in training or leave persistent residual error later. %\WJcomment{what do you mean ?} \WJcomment{I got lost.}
Existing RaQL theory characterizes the asymptotic requirements and links inner accuracy to outer convergence \citep[Theorem 4.7 and Remark 4.8]{huang2021}, but it does not prescribe an online finite-budget strategy for coordinating the two recursions.

%\WJcomment{do not understand what is trained reliably, too vague}
This paper asks: \emph{how can CVaR RaQL use a fixed transition-sample budget to reduce empirical Bellman residuals and mitigate the instability caused by fixed inner-loop settings.} %\WJcomment{they will definitely change}
We address this question with an adaptive training controller built around the original CVaR risk objective. The controller uses observed training quantities to adjust the inner step sizes, allocate transition samples across state--action cells, aggregate the sampled CVaR targets, and calibrate the outer learning-rate exponent, inner risk scale, and inner-loop depth. %Implementation details are introduced in the next section, and the numbered ablation variants are defined only in the experimental section. %\WJcomment{What is the Scheme 6 then ?}
%We likewise defer the numerical comparisons to the Results section, where the mean Q Bellman residual (MeanBEQ) %\WJcomment{full name ?}
%and mean value Bellman residual (MeanBEV) %\WJcomment{full name ?}
%are formally defined and reported.

Our contributions are: %\WJcomment{each point shall give very clear expression.}
\begin{itemize}
  \item %\WJcomment{?}
  We identify a CVaR-specific error-amplification mechanism: inaccurate tracking of the moving VaR estimate can magnify errors in the target used by the outer Q-update.
  \item We develop a three-module adaptive training controller that stabilizes inner risk estimation, reallocates samples as training progresses, and calibrates key training parameters during training process.
  \item We use cumulative ablations to measure the effect of each controller mechanism to assess sensitivity to the CVaR level, discount factor, and sampling budget. The adaptive controller substantially reduces empirical Bellman residuals relative to the fixed-parameter baseline.
  \item %\WJcomment{Add the contribution on the application...}
  We demonstrate the adaptive controller in a chronological Bitcoin trading study using market and sentiment states and accounting for turnover costs, showing better risk-adjusted performance and lower downside-risk measures than the fixed-parameter baseline.
\end{itemize}

\section{Adaptive Finite-Budget Controller}
%\WJcomment{for each of the module, not just write the technical stuff, also explain the principle and reason of designing such module.}
We keep the CVaR RaQL estimator fixed and modify only the training controller. The controller is organized into three modules and evaluated through six schemes. %Each scheme retains all mechanisms from the preceding scheme and adds one new mechanism. In other words, the schemes are cumulative rather than parallel.%\WJcomment{do not mention scheme here. nobody can catch the meaning}

\subsection{Adaptive CVaR RaQL}
\label{sec:final-training-algorithm}
The procedure of our final adaptive CVaR RaQL algorithm is presented in Algorithm~\ref{alg:adaptive-cvar-raql}. Its main mechanisms are described in detail later.

\begin{algorithm}[t]
\caption{Adaptive finite-budget CVaR RaQL}
\label{alg:adaptive-cvar-raql}
\begin{algorithmic}[1]
\REQUIRE Training samples $\mathcal{D}$;  $B$; $\alpha,\gamma$; $k_w,\kappa_h,k_T,p$
\STATE Set $Q(s,a)\gets0$, $y(s,a)\gets0$, $n_{sa}\gets1$, and $b\gets0$
\STATE Use one epoch to estimate $\ell_{\mathrm{avg}}$, $\ell_{\min}$, and $\ell_{\max}$; add its samples to $b$
\STATE $y_{\min}\gets x_{\min}\gets\ell_{\min}/(1-\gamma)$ and $y_{\max}\gets x_{\max}\gets\ell_{\max}/(1-\gamma)$
\STATE Set $\eta$, $h_y$, and $L$ using Equations~\eqref{eq:eta-calibration}, \eqref{eq:hy-calibration}, and~\eqref{eq:inner-depth}
\WHILE{$b<B$}
  \STATE Freeze $Q^{-}(s,a)\gets Q(s,a)$ and set $V^{-}(s)\gets\min_a Q^{-}(s,a)$
  \STATE Reset $k_{sa}\gets0$ and the queues $\mathcal{Y}_{sa},\mathcal{H}_{sa}\gets\varnothing$
  \FOR{$j=1,\ldots,\min\{L,B-b\}$}
    \STATE Observe $s$ and set $T\gets b/B$
    \STATE Select $a$ using Equation~\eqref{eq:phase-dependent-action-selector}
    \STATE Observe loss $\ell$ and next state $s'$; set $b\gets b+1$ and $k_{sa}\gets k_{sa}+1$
    \STATE Append $y(s,a)$ to $\mathcal{Y}_{sa}$
    \STATE Calculate $\bar y_{sa}$ by averaging the latter half of $\mathcal{Y}_{sa}$
    \STATE $x\gets\ell+\gamma V^{-}(s')$ and $\widehat q_{sa}\gets G_\alpha(x,\bar y_{sa})$
    \STATE Calculate $\lambda_{sa}$ by Equation~\eqref{eq:lambda-decay}
    \STATE $y(s,a)\gets\operatorname{clip}\!\left(y(s,a)-\lambda_{sa}h_y g_y(x,\bar y_{sa}),y_{\min},y_{\max}\right)$
    \IF{$T\leq0.05$}
      \STATE Correct $y(s,a)$ by Equation~\eqref{eq:y-correction}
    \ENDIF
    \STATE Append $\widehat q_{sa}$ to $\mathcal{H}_{sa}$
  \ENDFOR
  \FORALL{$(s,a)$ visited in this outer iteration}
    \STATE Calculate $\widehat q_{\mathrm{agg},sa}$ using Equation~\eqref{eq:progressive-suffix-aggregation}
    \STATE $Q(s,a)\gets(1-n_{sa}^{-\eta})Q^{-}(s,a)+n_{sa}^{-\eta}\widehat q_{\mathrm{agg},sa}$
    \STATE $n_{sa}\gets n_{sa}+1$
  \ENDFOR
\ENDWHILE
\RETURN $Q$ and its greedy policy
\end{algorithmic}
\end{algorithm}

In Algorithm~\ref{alg:adaptive-cvar-raql}, $B$ denotes the training budget, $b$ is the number of training samples already used, and $T=b/B\in[0,1]$ denotes the current training progress. The parameters $k_w$, $\kappa_h$, and $k_T$ are calibration hyperparameters introduced in Section~\ref{sec:observable-calibration}. The parameters $p$ and $\eta$ are two decay exponents. The parameter $h_y$ denotes the inner risk scale. It converts the inner subgradient $g_y$ into an appropriate step size for the $y$-update. The parameter $L$ denotes the inner-loop depth, while $j$ denotes the number of inner updates in the current outer iteration. For each state--action cell $(s,a)$, $\lambda_{sa}$ denotes the step size, $n_{sa}$ denotes the number of times that cell has been updated in the outer loop, and $k_{sa}$ denotes the number of times it has been visited in the inner loop of the current outer iteration. The queues $\mathcal{Y}_{sa}$ and $\mathcal{H}_{sa}$ store the sequences of $y(s,a)$ estimates and $\widehat q_{sa}$ estimates, respectively, in the current outer iteration.

\subsection{Tracking-Stability Module}

\paragraph{Inner-Loop Decay.}
\label{mech:inner-loop-decay}
The baseline decays the step size by a global inner-loop counter $j$. However, the per-cell inner-update counter $k_{sa}$, rather than the global inner-update counter $j$, better reflects the progress in estimating the minimizer for each state--action cell. In the baseline, if a state--action cell is sampled for the first time when $j$ is large, the step size used for its first update has already decayed significantly. This can destabilize training. We instead use
\begin{equation}
  \lambda_{sa}=k_{sa}^{-p},
\end{equation}
This can better match the step size to the progress in estimating the minimizer for each state--action cell. %\WJcomment{I cannot quite catch them, many statements are imprecise.}
%\JJmodified{Here, the baseline uses $\lambda_j=j^{-p}$ for the $j$-th inner transition, regardless of which cell is visited. If cell $(s,a)$ is first encountered at a large $j$, its first update is therefore already multiplied by the small factor $j^{-p}$. We replace $j$ with $k_{sa}$, the number of visits to that cell within the current outer iteration, so the first visit to every cell has $\lambda_{sa}=1$. This removes the late-arrival throttling of newly visited cells. Because $k_{sa}$ is usually small and resets after each outer iteration, the change alone also weakens late-stage annealing; consistently, Scheme 1 increases the terminal residual and is intended to be paired with the outer-count decay of $\lambda_{sa}$ introduced next.}

\paragraph{Outer-Loop Decay.}
\label{mech:outer-loop-decay}
The outer loop estimates the Q-table, while the inner loop tracks the minimizer of the risk subproblem. Each Q-table update shifts this minimizer. %\WJcomment{Can we use change or shift ?}
The inner loop must then track its new location. As the outer learning rate decays, later Q-updates become smaller and the minimizer changes more slowly. The inner step size $\lambda_{sa}$ should therefore decay with the outer-update count at the same rate as the Q-update step size. The baseline has no such outer-scale decay. For the same inner visit count ($k_{sa}$), a late-stage update (large $n_{sa}$) uses the same step size as an early-stage update (small $n_{sa}$). This can push an accurate estimate away from the current minimizer and continually introduce noise. We therefore use
\begin{equation}
  \lambda_{sa}=k_{sa}^{-p}n_{sa}^{-\eta},
  \label{eq:lambda-decay}
\end{equation}
where $\eta$ is matched to the outer Q-learning exponent. This couples the inner step size to the Q-update rate. %\JJmodified{When $n_{sa}$ is small early in training, the factor $n_{sa}^{-\eta}$ allows relatively large inner updates, helping the minimizer estimate track the rapidly changing Q-table. When $n_{sa}$ becomes large later in training, this factor makes the inner updates smaller because the Q-table and the corresponding minimizer change more slowly. This reduces unnecessary perturbations to a late period training.}
%\WJcomment{I do not understand it at all.}

\paragraph{Early $y$ Correction.}
\label{mech:early-y-correction}
As shown in Equation~\eqref{eq:amplification}, using a $y$ estimate that is far below its minimizer $y^\star$ in the Q update can amplify estimation errors and cause training to diverge. Early in training, an inappropriate inner risk scale $h_y$, an unsuitable step size $\lambda_{sa}$, an insufficient inner-loop depth $L$, or limited updates for a hard-to-reach state--action cell can leave $y$ far below its current minimizer $y^\star$. At this stage, the Q-table is initialized at zero and changes rapidly, causing $y^\star$ to shift rapidly as well. Consequently, $y$ may fail to track the moving minimizer $y^\star$. For CVaR, $y^\star$ is the VaR of $X(s,a)$. Thus, $y^\star$ and the sampled target $x(s,a)$ are on the same scale. We use the clipped $x(s,a)$ only as a reference and pull the early-stage $y$ estimate toward it:
\begin{equation}
  y \leftarrow y+
  \frac{1}{n_{sa}+2}0.5^{k_{sa}-1}
  \left(
  \operatorname{clip}(x,y_{\min},y_{\max})-y
  \right).
  \label{eq:y-correction}
\end{equation}
Because the clipped target is not an accurate estimate of $y^\star$ and the risk of divergence is greatest in early training, we apply the correction only during the first $5\%$ of the budget. Its strength also decays with both outer and inner visits ($n_{sa}, k_{sa}$). The correction therefore weakens as the estimate becomes more accurate and does not affect late-stage convergence.

\subsection{Sampling-Efficiency Module}

\paragraph{Two-Phase Action Selection.}
\label{mech:two-phase-action-selection}
As Equation~\eqref{eq:one-step-loss-target} shows, the estimate of a cell $(s,a)$ depends on the value $V_Q(s')$. Early in training, the greedy action that determines $V_Q(s')$ is not yet stable. Any action could eventually become the greedy action. Accurate estimation therefore requires sufficient training across a broad range of Q-table cells. Concentrating updates on certain cells in the early stage may also deteriorate the training performance. The outer visit counts $n_{sa}$ of these cells can grow rapidly while the corresponding next-state values $V_Q(s')$ remain inaccurate. Their step sizes may then decay excessively. As the rest of the Q- table becomes more accurate and the targets change, the Q-value update and minimizer estimates for these cells become difficult. In the late stage, the greedy actions are generally stable. At this stage, non-greedy cells do not propagate errors or determine state values. Their importance therefore decreases. The budget should then shift toward the greedy cells that determine $V_Q$. These observations motivate the following two-phase action selector:
%\WJcomment{What's $n(s,a)$ ?} \JJcomment{this should be the already defined count $n_{sa}$:}
\begin{align}
& a_{\mathrm{inner}}(s)= \nonumber
\\
& \begin{cases}
\arg\min_{a\in\mathcal A} n_{sa}, & T<0.60,\\
\epsilon\text{-greedy with greedy probability }0.90, & T\ge 0.60.
\end{cases}
\label{eq:phase-dependent-action-selector}
\end{align}

\paragraph{Progressive Suffix Aggregation.}
\label{mech:progressive-suffix-aggregation}
Let $\widehat q_{1,sa},\ldots,\widehat q_{k_{sa},sa}$ be the sequence of inner CVaR targets for a given $(s,a)$ in one outer iteration. The baseline uses only $\widehat q_{k_{sa},sa}$. We instead use a progress-dependent suffix average:
\begin{equation}
 \widehat q_{\mathrm{agg},sa}
  =
  \frac{1}{m}\sum_{i=k_{sa}-m+1}^{k_{sa}}\widehat q_{i,sa},
  \qquad
  m=\lceil \omega(T) k_{sa}\rceil,
  \label{eq:progressive-suffix-aggregation}
\end{equation}
where $\omega(T)=0.1+0.1\min\{9,\lfloor 10T\rfloor\}$. Early in training, the inner variables are biased and only the tail of the sequence is trusted; later, more estimates are averaged to reduce target variance.
As training progresses and the inner variable starts closer to its current minimizer, the expanding window reuses more of the targets already computed. This reduces target variance without allowing strongly transient early iterates to dominate the outer update, motivated by the variance-reduction principle of iterate averaging \citep{li2023}.

\subsection{Observable-Calibration Module}
\label{sec:observable-calibration}

To reduce manual tuning across different settings, the controller calibrates the learning-rate exponent($\eta$), inner risk scale, and inner-loop depth separately.

\paragraph{Discount-aware outer learning-rate decay.}
\label{mech:discount-aware-outer-decay}
The outer Q-learning step-size is $n_{sa}^{-\eta}$.
%A larger $\gamma$ \textcolor{teal}{extends the effective horizon}, so values \textcolor{teal}{farther\WJcomment{farther?} along the trajectory} remain relevant.
A larger $\gamma$ extends the effective horizon, so losses occurring further in the future retain greater weight in the current value estimate. The outer Q-learning step-size should therefore remain responsive for longer and decay more slowly as $\gamma$ increases. RaQL requires the exponent to lie in $(1/2,1]$~\cite{huang2021}. We set
\begin{equation}
  \eta
  =
  \operatorname{clip}\left(
  0.5+k_w(1-\gamma),\,0.5+\varepsilon,\,1.0
  \right).
  \label{eq:eta-calibration}
\end{equation}
As $\gamma$ increases, $\eta$ decreases and the outer Q-learning rate decays more slowly. The clipping keeps $\eta$ within the admissible RaQL range.

\paragraph{Loss-aware inner risk scale.}
\label{mech:loss-aware-inner-scale}
Early in training, $y$ must track the moving minimizer $y^\star$. We set the expected unattenuated one-step movement, $h_y\mathbb{E}[|g_y|]$, to a fixed fraction of the scale of $y$. This scale is approximately $\ell_{\mathrm{avg}}/(1-\gamma)$. Since $\mathbb{E}[|g_y|]=2\alpha$ at $y^\star$, we set
\begin{equation}
h_y
  =
  \kappa_h
  \frac{\ell_{\mathrm{avg}}}{\alpha(1-\gamma)}.
  \label{eq:hy-calibration}
\end{equation}
For $\Delta y=-\lambda_{sa}h_y g_y$, the resulting update also satisfies
\begin{equation}
  \begin{cases}
    \mathbb{E}[\Delta y\mid y]>0, & y<y^\star,\\
    \mathbb{E}[\Delta y\mid y]=0, & y=y^\star,\\
    \mathbb{E}[\Delta y\mid y]<0, & y>y^\star.
  \end{cases}
\end{equation}
The rule therefore improves early tracking without changing the convergence point $y^\star$. To estimate $\ell_{\mathrm{avg}}$, we use a short warmup before training.
Because the trading loss is signed, we compute $\ell_{\mathrm{avg}}$ as the mean absolute loss over the warmup epoch, rather than as the signed sample mean, so that it provides a positive magnitude scale. The identity $\mathbb{E}[|g_y|]=2\alpha$ is exact at $y^\star$ when the target distribution is continuous at the quantile and $\mathbb{P}(X>y^\star)=1-\alpha$. Here, it is used only for scale calibration, and the resulting rule is evaluated across the reported $\alpha$ sweep.

\paragraph{Budget-aware inner-loop depth.}
\label{mech:budget-aware-inner-depth}
Under a fixed budget, $L$ balances inner-target accuracy against the number of outer Q-learning updates. Too few inner transitions leave an inaccurate CVaR target, while too many reduce the available outer updates. Motivated by the cube-root budget-allocation rule for nested simulation in Gordy and Juneja~\cite{gordy2010}, we use the average per-cell budget $B/(SA)$ and set
\begin{equation}
  L
  =
  \operatorname{round}\left(
  k_T\left(\frac{B}{SA}\right)^{1/3}
  \right).
  \label{eq:inner-depth}
\end{equation}
Because the nested-simulation allocation rule is not derived specifically for RaQL, we use the cube-root expression as an observable budget-scaling heuristic rather than claim that it is theoretically optimal for this setting. Its practical behavior is evaluated in the reported budget sweep.
Together, these three rules form the observable-calibration module. They set the learning-rate exponent, inner risk scale, and inner-loop depth from known or observed quantities. The coefficients $k_w$, $\kappa_h$, and $k_T$ are selected once and kept fixed across the reported parameter sweeps.

\section{Cryptocurrency Trading Experiment}

\subsection{Data and Market Environment}

We construct a daily cryptocurrency dataset by aligning Bitcoin market data with sentiment information. The Crypto Fear and Greed Index is retrieved from the Alternative.me public API \citep{alternativeMeFng}; it ranges from 0 to 100 and is accompanied by a categorical label such as Extreme Fear, Fear, Neutral, Greed, or Extreme Greed. Daily BTCUSDT closing prices are retrieved from the Binance public market-data API \citep{binanceSpotApi}. From the closing-price series, we compute daily return as $r_t=P_t/P_{t-1}-1$, where $P_t$ is the BTCUSDT close on day $t$. We also construct a sentiment-momentum feature, defined as the seven-day change in the Fear and Greed Index. After aligning both sources by UTC calendar date and removing missing feature values, the dataset contains 3,059 daily observations from February 8, 2018 to June 28, 2026.

We use a chronological split: the first $70\%$ of observations form the training set and the remaining $30\%$ form the out-of-sample test set. The state space is obtained by discretizing market sentiment, sentiment momentum, and recent return into $3^3=27$ discrete states. The action set consists of six portfolio positions,
\[
  \mathcal{A}=\{-1,-0.6,-0.2,0.2,0.6,1\},
\]
where $w(a)$ denotes the BTC exposure associated with action $a$.

\subsection{Training Protocol}
During training, the generic immediate loss $\ell(s_t,a,s_{t+1})$ is instantiated as
\[
\ell_t(a)=-100\,w(a)r_{t+1}.
\]
The factor 100 rescales daily returns to a numerically convenient range. Transaction costs are not included in the tabular training loss because they depend on the previous position; including them would require augmenting the Markov state with the previous action. In the out-of-sample trading evaluation, however, we deduct turnover costs:
\begin{equation}
  R^{\mathrm{net}}_t
  =
  w(a_t)r_{t+1}
  -
  c_{\mathrm{tc}}|w(a_t)-w(a_{t-1})|.
\end{equation}
This makes the trading evaluation conservative relative to the training objective.

All main ablation experiments use 20 random seeds, 400 epochs, and a total training sample budget of $B$=856,000 inner-transition samples. One epoch consists of one pass through the 2,140 adjacent transitions in the training set. Thus, 400 epochs correspond to $B$=856,000 inner-transition samples, including the warmup samples. Multiple seeds reduce the risk that an apparent improvement is caused by training randomness or a favorable run \citep{henderson2018,agarwal2021}; paired ablations additionally isolate the incremental effect of each controller component. The baseline, Scheme 0, is the fixed-parameter two-timescale CVaR RaQL implementation in Algorithm \ref{alg:baseline-raql}. It uses an inner-loop depth of $L=80$ and an inner risk scale of $h_y=10$, both of which remain fixed throughout training.

Schemes~1--6 form a cumulative ablation sequence. Scheme~$r$ retains all mechanisms included in Scheme~$r-1$ and adds only the new mechanism listed in row~$r$ of Table~\ref{tab:ablation}. Every later scheme contains all mechanisms introduced in the earlier schemes. %\WJcomment{Say that the modules in the previous scheme are all contained in the later scheme.} 

\subsection{Evaluation Protocol and Metrics}

We evaluate fixed-point accuracy using a sample-based empirical CVaR Bellman residual on the training set. We do not estimate or use an explicit transition matrix for evaluation. For a learned table $Q$, define
\begin{equation}
  (\widehat{\mathcal{T}}_{\rho}Q)(s,a)
  =
  \widehat{\rho}_{\alpha}
  \left(
  \left\{
  \ell_t(a)+\gamma V_Q(s_{t+1})
  : t\in\mathcal{I}(s)
  \right\}
  \right),
  \label{eq:empirical-bellman}
\end{equation}
where $\mathcal{I}(s)$ is the set of training indices with state $s$ and $\widehat{\rho}_{\alpha}$ is the empirical CVaR operator
\begin{equation}
  \widehat{\rho}_{\alpha}(\{x_i\}_{i=1}^m)
  =
  \min_{y\in\mathbb{R}}
  \left\{
  y+\frac{1}{(1-\alpha)m}\sum_{i=1}^{m}(x_i-y)_+
  \right\}.
\end{equation}
We report Q-level and V-level residuals,
\begin{align}
  e_Q(s,a)&=\left|(\widehat{\mathcal{T}}_{\rho}Q)(s,a)-Q(s,a)\right|,\\
  e_V(s)&=\left|\min_a(\widehat{\mathcal{T}}_{\rho}Q)(s,a)-V_Q(s)\right|.
\end{align}
The tables focus on mean Bellman residuals, with maximum residuals used as a robustness diagnostic. We summarize these quantities using the mean empirical Q-level Bellman residual (MeanBEQ), maximum empirical Q-level Bellman residual (MaxBEQ), mean empirical value-level Bellman residual (MeanBEV), and maximum empirical value-level Bellman residual (MaxBEV).

For out-of-sample financial performance, the learned table induces the deterministic policy
\begin{equation}
  a_t=\arg\min_{a\in\mathcal{A}}Q(s_t,a).
\end{equation}
The policy is evaluated once on the 918 observations in the chronological test set, with no Q-updates on test observations. We use a transaction cost of $c_{\mathrm{tc}}=0.0005$ (5 basis points) and set the position immediately before the test period to zero. For each of the 20 trained tables, we report cumulative return, annualized return, annualized volatility, the Sharpe ratio with zero risk-free rate, maximum drawdown, average turnover, and empirical CVaR of daily loss. Scheme results are summarized by their mean and sample standard deviation across seeds. These standard deviations measure variation due to training randomness on the same market path. Buy-and-hold and cash/no-trade are deterministic baselines. We also include fixed-exposure policies with $w=0.2$ and $w=-0.2$ as deterministic baselines.

For daily net return $R_t^{\mathrm{net}}$, cumulative and annualized returns are
\begin{align}
  \mathrm{CumRet}
  &=\prod_{t=1}^{N}(1+R_t^{\mathrm{net}})-1,\\
  \mathrm{AnnRet}
  &=\left[\prod_{t=1}^{N}(1+R_t^{\mathrm{net}})\right]^{365}/N-1.
\end{align}
If $\bar R$ and $s_R$ are the sample mean and standard deviation of daily net returns, then
\begin{equation}
  \mathrm{AnnVol}=\sqrt{365}\,s_R,
  \qquad
  \mathrm{Sharpe}=\sqrt{365}\,\frac{\bar R}{s_R}.
\end{equation}
Thus, the reported annualized metrics use a 365-observation financial-market convention.Because Bitcoin trades continuously, this annualization uses 365 calendar-day observations.
Maximum drawdown is computed from the net-return wealth path, and average turnover is
\begin{equation}
  \mathrm{Turnover}
  =\frac{1}{N}\sum_{t=1}^{N}|w_t-w_{t-1}|.
\end{equation}
where $w_t:=w(a_t)$ denote the BTC exposure selected on day $t$. Finally, test CVaR is the empirical upper-tail CVaR of daily loss $L_t=-R_t^{\mathrm{net}}$ at $\alpha=0.6$; lower values indicate less severe tail loss.

\section{Results}

%\WJcomment{let's add some figures.}

\subsection{Ablation of Controller Components %\WJcomment{We call them module, right ?}
}
%\YFcomment{We have three modules; however, we have six \textcolor{teal}{components/mechanisms/schemes.}}

Table~\ref{tab:ablation} reports the ablation results at the end of training. The final adaptive controller, Scheme 6, reduces MeanBEQ %\WJcomment{MeanBEQ, right ? Match each full name with short name.}
from 1.2202 to 0.1854 and MeanBEV from 1.1624 to 0.0535 relative to the fixed-parameter baseline.

\begin{table}[t]
\centering
\caption{Ablation results at the 100\% checkpoint. Lower is better.}
\label{tab:ablation}
\scriptsize
\begin{tabular}{@{}llrrrr@{}}
\toprule
Scheme & Added mechanism & MaxBEQ & MeanBEQ & MaxBEV & MeanBEV \\
\midrule
0 & \hyperref[alg:baseline-raql]{Fixed baseline} & 4.6889 & 1.2202 & 1.9789 & 1.1624 \\
1 & \hyperref[mech:inner-loop-decay]{Inner-Loop Decay} & 6.6569 & 3.5297 & 4.7099 & 3.3499 \\
2 & \hyperref[mech:outer-loop-decay]{Outer-Loop Decay} & 3.9745 & 0.6122 & 0.3134 & 0.1211 \\
3 & \hyperref[mech:early-y-correction]{Early $y$ Correction} & 3.9438 & 0.6080 & 0.3133 & 0.1207 \\
4 & \hyperref[mech:two-phase-action-selection]{Two-Phase Action Selection} & 1.4916 & 0.2369 & 0.2819 & 0.0950 \\
5 & \hyperref[mech:progressive-suffix-aggregation]{Progressive Suffix Aggregation} & 1.4361 & 0.2251 & 0.1667 & 0.0601 \\
6 & \hyperref[sec:observable-calibration]{Observable-Calibration Module} & 1.3302 & 0.1854 & 0.1959 & 0.0535 \\
\bottomrule
\end{tabular}
\end{table}

The gains do not come from a single mechanism. The
\hyperref[mech:inner-loop-decay]{Inner-Loop Decay} alone reduces the effective decay relative to the global-counter
baseline. This leaves larger steps late in training and increases the
terminal residuals. Introducing
\hyperref[mech:outer-loop-decay]{Outer-Loop Decay}
substantially reduces all four residual metrics.
\hyperref[mech:two-phase-action-selection]{Two-Phase Action Selection}
and \hyperref[mech:progressive-suffix-aggregation]{Progressive Suffix Aggregation}
then improve sample efficiency and further reduce the residuals. The
\hyperref[sec:observable-calibration]{Observable-Calibration Module}
preserves this accuracy with fewer hand-set constants.
%\WJcomment{Cite the corresponding section or equation when necessary}
%\YFcomment{I have added a hyperlink to each mechanism name that links to its corresponding description.}

\subsection{\texorpdfstring{Mechanism-Specific Evaluation}%
{Mechanism-Specific Evaluation}}
\begin{figure*}[t]
  \centering
  \includegraphics[scale=1]{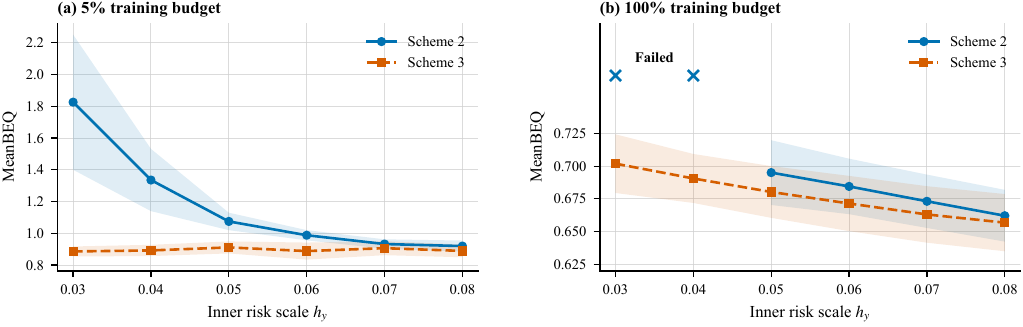}
  \caption{Stability across inner risk scales.}
  \Description{Two line charts compare Schemes 2 and 3 as the
  inner risk scale $h_y$ varies from 0.03 to 0.08. The left panel reports MeanBEQ
  at 5 percent of the training budget. The right panel reports terminal
  MeanBEQ; Scheme 2 fails at 0.03 and 0.04, while Scheme 3 succeeds at
  all six tested scales.}
  \label{fig:scheme2-vs3}
\end{figure*}

%\WJcomment{Again, link each scheme with corresponding section/equation}
\hyperref[mech:early-y-correction]{Scheme 3} introduces the early $y$ correction to improve
training stability and reduce failures. A training is considered failed if its MeanBEQ diverges. Figure~\ref{fig:scheme2-vs3}
presents the stress-test results for \hyperref[mech:outer-loop-decay]{Scheme 2} and \hyperref[mech:early-y-correction]{Scheme 3} across a range of
small inner risk scales ($h_y$). At
$h_y=0.03$ and $0.04$, all \hyperref[mech:outer-loop-decay]{Scheme 2} seeds fail.%\YFcomment{I put the explanation of failure in the second sentence of this paragraph.}
%\WJcomment{Fail means diverge in MeanBEQ value, right?} 
\hyperref[mech:early-y-correction]{Scheme 3}
succeeds for all seeds at every tested scale and keeps the 5\%-budget
MeanBEQ between 0.89 and 0.91. The correction therefore improves both
training stability and early estimation accuracy, with clearer benefits under stressful settings, particularly when $h_y$ is small.%\WJcomment{What's the explicit stressful setting.}

\begin{figure}[t]
  \centering
  \includegraphics[scale=1]{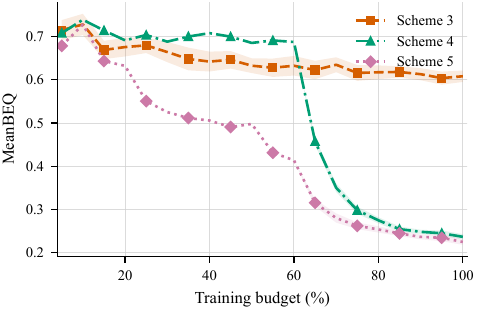}
  \caption{Mean Q residual during training.}
  \Description{A line chart compares the MeanBEQ trajectories of Schemes
  3, 4, and 5 over the training budget.}
  \label{fig:scheme3-to5}
\end{figure}

\hyperref[mech:two-phase-action-selection]{Scheme 4} and \hyperref[mech:progressive-suffix-aggregation]{Scheme 5} improve sample efficiency.
Figure~\ref{fig:scheme3-to5} presents their MeanBEQ trajectories together
with \hyperref[mech:early-y-correction]{Scheme 3}. In \hyperref[mech:two-phase-action-selection]{Scheme 4}, balanced coverage during the first 60\%
builds broad Q-table accuracy and avoids excessive step-size decay in
individual cells. Once action selection shifts toward greedy actions,
MeanBEQ drops from 0.69 at 60\% to 0.46 at 65\%. \hyperref[mech:progressive-suffix-aggregation]{Scheme 5} uses
progressive suffix aggregation to reduce sampling variance. It holds the
lowest MeanBEQ at nearly every checkpoint and achieves the highest sample
efficiency.

\begin{figure}[t]
  \centering
  \includegraphics[scale=1]{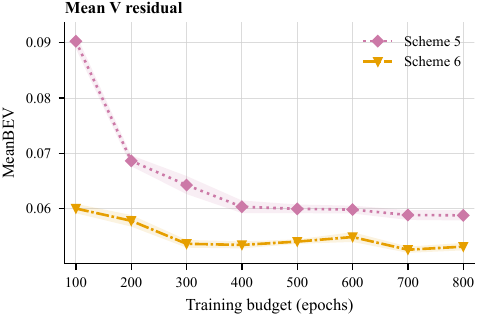}
  \caption{Mean residuals across training budgets.}
  \Description{Two line charts compare Schemes 5 and 6 over training
  budgets from 100 to 800 epochs. The left panel reports MeanBEQ and the
  right panel reports MeanBEV. Shaded bands are 95 percent confidence
  intervals over 10 seeds.}
  \label{fig:scheme5-vs6-budget}
\end{figure}

\hyperref[sec:observable-calibration]{Scheme 6} is designed to adapt its training parameters when
the risk level $\alpha$, discount factor $\gamma$, or training budget
changes. Table~\ref{tab:robustness} compares \hyperref[mech:progressive-suffix-aggregation]{Scheme 5} and \hyperref[sec:observable-calibration]{Scheme 6} across the
complete $\alpha$ and $\gamma$ sweeps. \hyperref[sec:observable-calibration]{Scheme 6} has lower MeanBEQ and
MeanBEV at every tested setting. It is also less sensitive to changes
in $\alpha$, while both schemes remain stable across $\gamma$.
Figure~\ref{fig:scheme5-vs6-budget} reports the budget sweep. \hyperref[sec:observable-calibration]{Scheme 6}
achieves lower MeanBEV than \hyperref[mech:progressive-suffix-aggregation]{Scheme 5} at every tested budget.
As the training budget decreases, \hyperref[sec:observable-calibration]{Scheme 6} remains more stable than
\hyperref[mech:progressive-suffix-aggregation]{Scheme 5}. Overall, \hyperref[sec:observable-calibration]{Scheme 6} performs better across all tested settings.

\input{table_scheme5_vs6_sensitivity.tex}

\subsection{Out-of-Sample Trading Performance}

Table~\ref{tab:trading} reports performance on the chronological test set after transaction costs. Scheme 6 achieves the highest risk-adjusted performance among all evaluated policies. Its mean Sharpe ratio is 0.9281, compared with 0.5628 for Scheme 0 and 0.4902 for buy-and-hold. It also lowers annualized volatility from 14.53\% for Scheme 0 to 9.57\%, maximum drawdown from 17.77\% to 6.46\%, and daily CVaR loss from 0.0050 to 0.0041. Scheme 6 also has much lower volatility, drawdown, and CVaR loss than buy-and-hold.

\begin{table*}[t]
\centering
\caption{Out-of-sample trading performance after transaction costs. Scheme entries are mean (sample standard deviation) across 20 seeds; values are rounded to four decimals.}
\label{tab:trading}
\scriptsize
\resizebox{\textwidth}{!}{%
\begin{tabular}{@{}lrrrrrrr@{}}
\toprule
Policy & CumRet & AnnRet & AnnVol & Sharpe & MaxDD & Turnover & CVaR loss \\
\midrule
Scheme 6 & 0.2361 (0.0118) & 0.0879 (0.0041) & 0.0957 (0.0000) & 0.9281 (0.0401) & 0.0646 (0.0038) & 0.1740 (0.0016) & 0.0041 (0.0000) \\
Scheme 0 & 0.1922 (0.1121) & 0.0714 (0.0389) & 0.1453 (0.0482) & 0.5628 (0.2281) & 0.1777 (0.0648) & 0.1278 (0.0577) & 0.0050 (0.0007) \\
Fixed $w=0.2$ & 0.1125 & 0.0433 & 0.0959 & 0.4902 & 0.1261 & 0.0002 & 0.0042 \\
Fixed $w=-0.2$ & -0.1219 & -0.0504 & 0.0959 & -0.4910 & 0.2415 & 0.0002 & 0.0045 \\
Buy-and-hold & 0.3543 & 0.1282 & 0.4793 & 0.4902 & 0.5221 & 0.0011 & 0.0210 \\
Cash/no-trade & 0.0000 & 0.0000 & 0.0000 & 0.0000 & 0.0000 & 0.0000 & 0.0000 \\
\bottomrule
\end{tabular}
}
\end{table*}

Scheme 6 earns a lower cumulative return than buy-and-hold (23.61\% versus 35.43\%), so the result does not indicate dominance in raw return. Instead, it shows a different risk-return profile: Scheme 6 accepts less market exposure and gives up part of the upside while reducing annualized volatility from 47.93\% to 9.57\% and maximum drawdown from 52.21\% to 6.46\%. Compared with the fixed $w=0.2$ baseline, Scheme 6 achieves a higher cumulative return (23.61\% versus 11.25\%). It also has a lower maximum drawdown (6.46\% versus 12.61\%). These results show that maintaining a low market exposure alone does not reproduce Scheme 6's improved risk--return profile. Scheme 6 is also more stable across seeds than Scheme 0. The standard deviations of its cumulative return, Sharpe ratio, and maximum drawdown are 0.0118, 0.0401, and 0.0038, respectively, compared with 0.1121, 0.2281, and 0.0648 for Scheme 0. Overall, the adaptive controller achieves smaller Bellman residuals, reduces variation in out-of-sample performance across training seeds, and lowers volatility, maximum drawdown, and CVaR loss.
\section{Discussion, Limitations, and Conclusion}

This paper studies a finite-budget instability in CVaR RaQL caused by three interacting factors: (i) the asymptotic convergence theory prescribes no rule for allocating a fixed sampling budget between inner and outer recursions, or across state–action cells with unequal visitation; (ii) inner and outer estimation errors are coupled through bootstrapping and can amplify under fixed hyperparameters; and (iii) uniform inner-loop configurations ignore per-cell visitation asymmetry, leaving some cells over-trained and others under-trained. The proposed adaptive controller addresses all three without changing the CVaR estimator or Bellman fixed point: per-cell step sizing and rate-matched decay counter the coupling, coverage-first-then-greedy allocation counters the asymmetry, and observable calibration replaces hand-tuned constants for portability. Under 856,000 inner-transition samples, the controller reduces MeanBEQ from 1.2202 to 0.1854 and MeanBEV from 1.1624 to 0.0535; the cumulative ablation confirms that no single mechanism dominates, consistent with the three-factor diagnosis. On the chronological test set, the Sharpe ratio rises from 0.5628 to 0.9281 and maximum drawdown drops from 17.77\% to 6.46\%; the return remains below buy-and-hold, confirming improved risk-adjusted rather than raw-return performance.

The main limitations of this work are: the controller lacks convergence guarantees for the coupled adaptive recursions, and the evaluation is confined to one asset with tabular states. Future work should develop this theory, extend to multi-asset and function-approximation settings, and examine whether the factor decomposition generalizes to other risk-aware RL algorithms.

\bibliographystyle{ACM-Reference-Format}
\bibliography{main}

\end{document}

%% file: table_scheme5_vs6_sensitivity.tex
\begin{table}[t]
\centering
\caption{Mean residual ranges for Schemes 5 and 6 across the $\alpha$ and $\gamma$ sweeps.}
\label{tab:robustness}
\scriptsize
\begin{tabular}{@{}llcc@{}}
\toprule
Sweep & Method & MeanBEQ range & MeanBEV range \\
\midrule
\multirow{2}{*}{$\alpha\in[0.50,0.90]$, $\gamma=0.80$} & Scheme 5 & 0.1910--0.6702 & 0.0528--0.1997 \\
 & Scheme 6 & 0.1663--0.4005 & 0.0500--0.1108 \\
\midrule
\multirow{2}{*}{$\gamma\in[0.70,0.90]$, $\alpha=0.60$} & Scheme 5 & 0.2224--0.2269 & 0.0594--0.0641 \\
 & Scheme 6 & 0.1856--0.1940 & 0.0520--0.0576 \\
\bottomrule
\end{tabular}
\end{table}

%% file: main.bib
@article{huang2021,
  author  = {Huang, Wenjie and Haskell, William B.},
  title   = {Stochastic Approximation for Risk-Aware Markov Decision Processes},
  journal = {IEEE Transactions on Automatic Control},
  year    = {2021},
  volume  = {66},
  number  = {3},
  pages   = {1314--1320}
}

@article{rockafellar2000,
  author  = {Rockafellar, R. Tyrrell and Uryasev, Stanislav},
  title   = {Optimization of Conditional Value-at-Risk},
  journal = {Journal of Risk},
  year    = {2000},
  volume  = {2},
  number  = {3},
  pages   = {21--41},
  doi     = {10.21314/JOR.2000.038}
}

@article{bental2007,
  author  = {Ben-Tal, Aharon and Teboulle, Marc},
  title   = {An Old-New Concept of Convex Risk Measures: The Optimized Certainty Equivalent},
  journal = {Mathematical Finance},
  year    = {2007},
  volume  = {17},
  number  = {3},
  pages   = {449--476}
}

@inproceedings{chow2014,
  author    = {Chow, Yinlam and Ghavamzadeh, Mohammad},
  title     = {Algorithms for {CVaR} Optimization in {MDP}s},
  booktitle = {Advances in Neural Information Processing Systems},
  year      = {2014},
  volume    = {27}
}

@article{watkins1992,
  author  = {Watkins, Christopher J. C. H. and Dayan, Peter},
  title   = {Q-Learning},
  journal = {Machine Learning},
  year    = {1992},
  volume  = {8},
  number  = {3--4},
  pages   = {279--292}
}

@article{borkar2000,
  author  = {Borkar, Vivek S. and Meyn, Sean P.},
  title   = {The {O.D.E.} Method for Convergence of Stochastic Approximation and Reinforcement Learning},
  journal = {SIAM Journal on Control and Optimization},
  year    = {2000},
  volume  = {38},
  number  = {2},
  pages   = {447--469}
}

@article{evendar2003,
  author  = {Even-Dar, Eyal and Mansour, Yishay},
  title   = {Learning Rates for {Q}-Learning},
  journal = {Journal of Machine Learning Research},
  year    = {2003},
  volume  = {5},
  pages   = {1--25}
}

@article{gordy2010,
  author  = {Gordy, Michael B. and Juneja, Sandeep},
  title   = {Nested Simulation in Portfolio Risk Measurement},
  journal = {Management Science},
  year    = {2010},
  volume  = {56},
  number  = {10},
  pages   = {1833--1848}
}

@inproceedings{bellemare2016,
  author    = {Bellemare, Marc G. and Srinivasan, Sriram and Ostrovski, Georg and Schaul, Tom and Saxton, David and Munos, R{\'e}mi},
  title     = {Unifying Count-Based Exploration and Intrinsic Motivation},
  booktitle = {Advances in Neural Information Processing Systems},
  year      = {2016},
  volume    = {29}
}

@inproceedings{tokic2010,
  author    = {Tokic, Michel},
  title     = {Adaptive {$\epsilon$}-Greedy Exploration in Reinforcement Learning Based on Value Differences},
  booktitle = {KI 2010: Advances in Artificial Intelligence},
  year      = {2010},
  series    = {Lecture Notes in Computer Science},
  volume    = {6359},
  pages     = {203--210},
  publisher = {Springer}
}

@inproceedings{li2023,
  author    = {Li, Xiang and Yang, Wenhao and Liang, Jiadong and Zhang, Zhihua and Jordan, Michael I.},
  title     = {A Statistical Analysis of Polyak--Ruppert Averaged {Q}-Learning},
  booktitle = {Proceedings of the 26th International Conference on Artificial Intelligence and Statistics},
  year      = {2023},
  series    = {Proceedings of Machine Learning Research},
  volume    = {206},
  pages     = {2207--2261}
}

@inproceedings{henderson2018,
  author    = {Henderson, Peter and Islam, Riashat and Bachman, Philip and Pineau, Joelle and Precup, Doina and Meger, David},
  title     = {Deep Reinforcement Learning That Matters},
  booktitle = {Proceedings of the AAAI Conference on Artificial Intelligence},
  year      = {2018},
  volume    = {32},
  number    = {1}
}

@inproceedings{agarwal2021,
  author    = {Agarwal, Rishabh and Schwarzer, Max and Castro, Pablo Samuel and Courville, Aaron and Bellemare, Marc G.},
  title     = {Deep Reinforcement Learning at the Edge of the Statistical Precipice},
  booktitle = {Advances in Neural Information Processing Systems},
  year      = {2021},
  volume    = {34},
  pages     = {29304--29320}
}

@misc{alternativeMeFng,
  author       = {{Alternative.me}},
  title        = {Crypto Fear \& Greed Index},
  year         = {n.d.},
  howpublished = {\url{https://alternative.me/crypto/fear-and-greed-index/}},
  note         = {Accessed 26 July 2026}
}

@misc{binanceSpotApi,
  author       = {{Binance}},
  title        = {Binance Spot REST API: Kline/Candlestick Data},
  year         = {n.d.},
  howpublished = {\url{https://developers.binance.com/en/docs/catalog/core-trading-spot-trading/api/rest-api/market\#klinecandlestick-data}},
  note         = {Accessed 26 July 2026}
}

@article{artzner1999,
  author  = {Artzner, Philippe and Delbaen, Freddy and Eber, Jean-Marc and Heath, David},
  title   = {Coherent Measures of Risk},
  journal = {Mathematical Finance},
  year    = {1999},
  volume  = {9},
  number  = {3},
  pages   = {203--228}
}

@article{acerbi2002,
  author  = {Acerbi, Carlo},
  title   = {Spectral Measures of Risk: A Coherent Representation of Subjective Risk Aversion},
  journal = {Journal of Banking \& Finance},
  year    = {2002},
  volume  = {26},
  number  = {7},
  pages   = {1505--1518}
}

@inproceedings{prashanth2014,
  author    = {Prashanth, L. A.},
  title     = {Policy Gradients for {CVaR}-Constrained {MDP}s},
  booktitle = {Algorithmic Learning Theory},
  year      = {2014},
  series    = {Lecture Notes in Computer Science},
  volume    = {8776},
  pages     = {155--169},
  publisher = {Springer}
}

@inproceedings{dabney2018,
  author    = {Dabney, Will and Rowland, Mark and Bellemare, Marc G. and Munos, R{\'e}mi},
  title     = {Distributional Reinforcement Learning with Quantile Regression},
  booktitle = {Proceedings of the AAAI Conference on Artificial Intelligence},
  year      = {2018},
  volume    = {32},
  number    = {1},
  doi       = {10.1609/aaai.v32i1.11791}
}

@inproceedings{lim2022,
  author    = {Lim, Shiau Hong and Malik, Ilyas},
  title     = {Distributional Reinforcement Learning for Risk-Sensitive Policies},
  booktitle = {Advances in Neural Information Processing Systems},
  year      = {2022},
  volume    = {35}
}

@inproceedings{chen2024,
  author    = {Chen, Yu and Zhang, Xiangcheng and Wang, Siwei and Huang, Longbo},
  title     = {Provable Risk-Sensitive Distributional Reinforcement Learning with General Function Approximation},
  booktitle = {Proceedings of the 41st International Conference on Machine Learning},
  year      = {2024},
  series    = {Proceedings of Machine Learning Research},
  volume    = {235},
  pages     = {7748--7791},
  publisher = {PMLR}
}

@inproceedings{konda2000,
  author    = {Konda, Vijay R. and Tsitsiklis, John N.},
  title     = {Actor-Critic Algorithms},
  booktitle = {Advances in Neural Information Processing Systems},
  year      = {1999},
  volume    = {12},
  pages     = {1008--1014}
}

@inproceedings{jin2018,
  author    = {Jin, Chi and Allen-Zhu, Zeyuan and Bubeck, S{\'e}bastien and Jordan, Michael I.},
  title     = {Is {Q}-Learning Provably Efficient?},
  booktitle = {Advances in Neural Information Processing Systems},
  year      = {2018},
  volume    = {31}
}

@article{duchi2011,
  author  = {Duchi, John and Hazan, Elad and Singer, Yoram},
  title   = {Adaptive Subgradient Methods for Online Learning and Stochastic Optimization},
  journal = {Journal of Machine Learning Research},
  year    = {2011},
  volume  = {12},
  pages   = {2121--2159}
}

@inproceedings{vanhasselt2016,
  author    = {van Hasselt, Hado and Guez, Arthur and Hessel, Matteo and Mnih, Volodymyr and Silver, David},
  title     = {Learning Values across Many Orders of Magnitude},
  booktitle = {Advances in Neural Information Processing Systems},
  year      = {2016},
  volume    = {29}
}

@misc{schaul2021,
  author       = {Schaul, Tom and Ostrovski, Georg and Kemaev, Iurii and Borsa, Diana},
  title        = {Return-Based Scaling: Yet Another Normalisation Trick for Deep {RL}},
  year         = {2021},
  howpublished = {arXiv preprint arXiv:2105.05347}
}

@article{moore1993,
  author  = {Moore, Andrew W. and Atkeson, Christopher G.},
  title   = {Prioritized Sweeping: Reinforcement Learning with Less Data and Less Time},
  journal = {Machine Learning},
  year    = {1993},
  volume  = {13},
  number  = {1},
  pages   = {103--130}
}

@inproceedings{schaul2016,
  author    = {Schaul, Tom and Quan, John and Antonoglou, Ioannis and Silver, David},
  title     = {Prioritized Experience Replay},
  booktitle = {International Conference on Learning Representations},
  year      = {2016}
}

@article{wang2022risk,
  title={Risk-averse autonomous systems: A brief history and recent developments from the perspective of optimal control},
  author={Wang, Yuheng and Chapman, Margaret P},
  journal={Artificial Intelligence},
  volume={311},
  pages={103743},
  year={2022},
  publisher={Elsevier}
}

@inproceedings{shen2014riskaverse,
  author    = {Shen, Yun and Huang, Ruihong and Yan, Chang and Obermayer, Klaus},
  title     = {Risk-Averse Reinforcement Learning for Algorithmic Trading},
  booktitle = {2014 IEEE Conference on Computational Intelligence for
               Financial Engineering \& Economics (CIFEr)},
  year      = {2014},
  pages     = {391--398},
  publisher = {IEEE},
  doi       = {10.1109/CIFEr.2014.6924100}
}

@inproceedings{bisi2020fx,
  author    = {Bisi, Lorenzo and Liotet, Pierre and Sabbioni, Luca and
               Reho, Gianmarco and Montali, Nico and Restelli, Marcello and
               Corno, Cristiana},
  title     = {Foreign Exchange Trading: A Risk-Averse Batch Reinforcement
               Learning Approach},
  booktitle = {Proceedings of the First ACM International Conference on
               AI in Finance},
  year      = {2020},
  articleno = {26},
  numpages  = {8},
  publisher = {Association for Computing Machinery},
  doi       = {10.1145/3383455.3422571}
}

@inproceedings{vittori2020hedging,
  author    = {Vittori, Edoardo and Trapletti, Michele and Restelli, Marcello},
  title     = {Option Hedging with Risk Averse Reinforcement Learning},
  booktitle = {Proceedings of the First ACM International Conference on
               AI in Finance},
  year      = {2020},
  articleno = {27},
  numpages  = {8},
  publisher = {Association for Computing Machinery},
  doi       = {10.1145/3383455.3422532}
}

@inproceedings{vadori2020martingale,
  author    = {Vadori, Nelson and Ganesh, Sumitra and Reddy, Prashant P. and
               Veloso, Manuela},
  title     = {Risk-Sensitive Reinforcement Learning: A Martingale Approach
               to Reward Uncertainty},
  booktitle = {Proceedings of the First ACM International Conference on
               AI in Finance},
  year      = {2020},
  articleno = {28},
  numpages  = {9},
  publisher = {Association for Computing Machinery},
  doi       = {10.1145/3383455.3422519}
}

@inproceedings{alameer2022cvar,
  author    = {Alameer, Ali and Alshehri, Khaled},
  title     = {Conditional Value-at-Risk for Quantitative Trading:
               A Direct Reinforcement Learning Approach},
  booktitle = {2022 IEEE Conference on Control Technology and Applications
               (CCTA)},
  year      = {2022},
  pages     = {1208--1213},
  publisher = {IEEE},
  doi       = {10.1109/CCTA49430.2022.9966017}
}

@article{cui2023crypto,
  author  = {Cui, Tianxiang and Ding, Shusheng and Jin, Huan and
             Zhang, Yongmin},
  title   = {Portfolio Constructions in Cryptocurrency Market:
             A {CVaR}-Based Deep Reinforcement Learning Approach},
  journal = {Economic Modelling},
  year    = {2023},
  volume  = {119},
  pages   = {106078},
  doi     = {10.1016/j.econmod.2022.106078}
}

@article{coache2023dynamic,
  author  = {Coache, Anthony and Jaimungal, Sebastian and
             Cartea, {\'A}lvaro},
  title   = {Conditionally Elicitable Dynamic Risk Measures for Deep
             Reinforcement Learning},
  journal = {SIAM Journal on Financial Mathematics},
  year    = {2023},
  volume  = {14},
  number  = {4},
  pages   = {1249--1289},
  doi     = {10.1137/22M1527209}
}

@article{enkhsaikhan2024portfolio,
  author  = {Enkhsaikhan, Bayaraa and Jo, Ohyun},
  title   = {Risk-Averse Reinforcement Learning for Portfolio Optimization},
  journal = {ICT Express},
  year    = {2024},
  volume  = {10},
  number  = {4},
  pages   = {857--862},
  doi     = {10.1016/j.icte.2024.04.010}
}
